\documentclass{article}

\usepackage{speed}

\usepackage[utf8]{inputenc} % allow utf-8 input
\usepackage[T1]{fontenc}    % use 8-bit T1 fonts
\usepackage{hyperref}       % hyperlinks
\usepackage{url}            % simple URL typesetting
\usepackage{booktabs}       % professional-quality tables
\usepackage{amsfonts}       % blackboard math symbols
\usepackage{nicefrac}       % compact symbols for 1/2, etc.
\usepackage{microtype}      % microtypography
\usepackage{xcolor}         % colors
\usepackage{graphicx}       % graphics
\graphicspath{ {images/} } 
\usepackage{tabulary}
\usepackage{caption}
\usepackage{subcaption}
\usepackage{float}
\usepackage{amsmath}
\usepackage{multirow}

\title{Ego Vehicle Speed Estimation using 3D Convolution with Masked Attention}

\author{ {\hspace{1mm}Athul M. Mathew} \\
	Elm Company\\
	Saudi Arabia \\
	\texttt{amathew@elm.sa} \\
	\And
	{\hspace{1mm}Thariq Khalid} \\
	Elm Company\\
	Saudi Arabia \\
	\texttt{tkadavil@elm.sa} \\
}

\def\plainkeywords{: 3D-CNN; masked-attention ; ego-vehicle ; speed estimation; video transformer; KITTI; nuImages}

\hypersetup{%
  pdfkeywords={\plainkeywords},
}
\begin{document}

\maketitle

\begin{abstract}
Speed estimation of an ego vehicle is crucial to enable autonomous driving and advanced driver assistance technologies. Due to functional and legacy issues, conventional methods depend on in-car sensors to extract vehicle speed through the Controller Area Network (CAN) bus. However, it is desirable to have modular systems that are not susceptible  to external sensors to execute perception tasks.  
In this paper, we propose a novel 3D-CNN with masked-attention (3DCMA) architecture to estimate ego vehicle speed using a single front-facing monocular camera. To demonstrate the effectiveness of our method, we conduct experiments on two publicly available datasets, nuImages and KITTI. Our method outperforms the current state-of-the-art architecture for video vision by 27\% and 34\% in nuImages and KITTI datasets, respectively. We also demonstrate masked-attention's efficacy by comparing our method with a traditional 3D-CNN. Our method achieved an error reduction of 23\% and 25\% for the datasets mentioned above when compared against 3D-CNN without masked-attention.
\end{abstract}


\section{INTRODUCTION}
The impact of electric vehicles today in contributing to an energy-efficient and sustainable world is immense\cite{hill2019role}. It is a significant influencing factor in the global push against climate change. To this end, self-driving vehicles add further value in enabling smart mobility, planning, and control for intelligent transportation systems. According to \cite{sun2014velocity}, predicting the ego vehicle speed reduces fuel consumption and optimizes cruise control \cite{stanger2013model}. Ego vehicle speed estimation is one of the fundamental steps toward Advanced Driver Assistance Systems(ADAS). Self-driving cars use stereo cameras, LiDAR (Light Detection and Ranging), and radars to estimate the speed of the ego vehicle and other vehicles. The combination of these sensors gives great accuracy leading to meticulous navigation to avoid crashes and ensure the safety of the vehicle and its surrounding. Camera-LiDAR fusion exploits the video stream and 3D point clouds simultaneously to get depth information of objects around the vehicle. Thus, the speed can be estimated using the multitude of features learned using deep learning. Visual Odometry(VO) of self-driving vehicles helps in ego-motion and thus helps in localization. Yet another way of obtaining the speed information is through the in-car sensors through the CAN bus. Estimating the speed of vehicles on roads has a vital role in traffic control and planning. The speed of vehicles on the road gives the state of the traffic environment, which will help forecast traffic congestion and monitor any anomalies. Autonomous vehicles use speed estimation to help themselves in navigation and mapping. Speed estimation could also find great value in Vehicle to Vehicle (V2V) applications for communication between vehicles on the road. 

\begin{figure}[h]
\centering
\includegraphics[width=160mm]{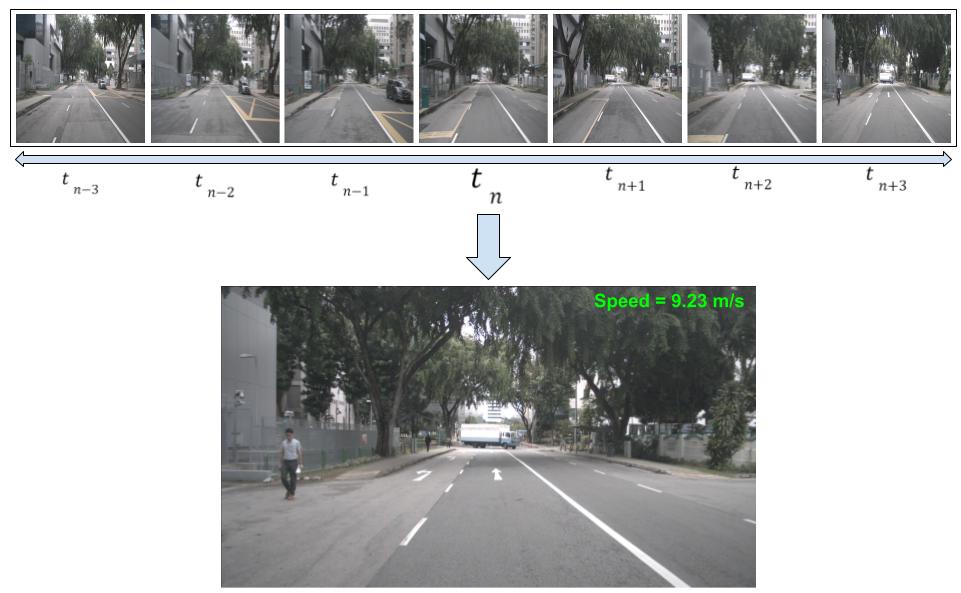} 
\caption{Estimation of ego-vehicle speed (\textcolor{green}{green}) using a continuous camera stream}
\label{fig:yolop}
\end{figure}

So far, very little work has been done using monocular camera-based speed estimation from a moving car. In our work, we present a 3D Convolutional Neural Network (3D-CNN) architecture trained on short videos using their grayscale image frames and the corresponding lane line segmentation masks. Using our neural network architecture, we are able to estimate the speed of the ego vehicle, which can, in turn help to estimate the speed of vehicles of interest (VOI) in the surrounding environment.

In the remaining sections of this paper, we present the related work, including existing speed estimation techniques using cameras and other sensors. We explain the network architecture in detail within the methods section, followed by the experiments section that describes the datasets incorporated for this work. In the last part of the paper, we discuss our findings and results and propose future work that we believe could contribute to this research. We believe that our work is effective yet simple and can be helpful as modular components in autonomous or intelligent traffic systems.

\section{RELATED WORK}
\subsection{Speed Estimation}
The work done by \cite{yamaguchi2006vehicle} is one of the early works to estimate the ego-motion using correspondence points detection, road region detection, moving object detection, and other derived features. Furthermore, 8-point algorithm\cite{hartley1997defense} and RANSAC\cite{fischler1981random} are applied to get the essential matrix of ego-motion. The  recent work in \cite{bandari2021end} implemented an end-to-end CNN-LSTM network to estimate the speed of an ego vehicle. The work performs evaluation on DBNet\cite{8578713} and comma.ai speed challenge dataset\cite{commaspeedchallenge}.
Other works, such as \cite{mejia2021vehicle}, propose speed estimation of vehicles from a CCTV point of view. Most require camera calibration and fixed view so that the vehicles pass through certain lines or regions of interest.

FlowNet\cite{ilg2017flownet} and PWC-Net\cite{sun2018pwc} are deep neural networks to estimate optical flow in videos. Further research in\cite{rill2019speed,hayakawa2019ego} make use of FlowNet or PWC-Net to estimate the ego vehicle speed. However, they perform ego vehicle speed estimation by further post-processing on the optical flow pixel velocity. None of the works demonstrate the end-to-end architecture capability where the speed could be learned with differentiation of the loss function.

\subsection{3D Convolutional Neural Network}
2D Convolutional Neural Networks have proven to be excellent at extracting feature maps for images. They are predominantly used for understanding the spatial aspects of images relevant to classification and object detection. However, they cannot capture the spatio-temporal features of videos spread across multiple continuous frames. 3D Convolutional Neural Networks are the best in learning spatio-temporal features and thus help in video classification \cite{karpathy2014large}, human action recognition \cite{ji20123d}, and sign language recognition \cite{pu2019iterative}. There have been recent works \cite{girdhar2019video,wu2019long,wang2018non} which use attention on top of 3D-CNN; however, they are limited to action recognition use cases. Very few works such as \cite{grinciunaite2016human, deng2017hand3d, ge20173d} perform regression using 3D-CNNs however they perform spatial localization-related tasks such as human pose or 3D hand pose. Our work performs the regression across the spatio-temporal aspects by having the 3D-CNN attend to both the visual features of the gray images and the lane masks.

\subsection{Vision Transformers}
Vision Transformers(ViTs) capitalized on the success of Transformers\cite{vaswani2017attention} in the field of Natural Language Processing. The basic Vision Transformer\cite{dosovitskiy2020image} takes non-overlapping patches of an image and creates token embeddings after performing linear projection. These embeddings are concatenated with position embeddings, after which they go through the transformer block, which contains layer normalization, Multi-Head Attention, and MLP operations to produce a classification output finally. Although the ViTs claim to replace the CNNs, the former lack the inductive bias, whereas the latter is translation invariant as well due to the local neighborhood structure of the convolution kernels. Moreover, transformers have quadratic complexity for their operations and scales with the input dimensions. The main advantage of ViTs is the global attention and long-range interaction. It is interesting to note that the hybrid CNN-Transformer with a CNN backbone outperforms the pure ViT approach.

Transformer models for video have been proposed in a plethora of architectures. The Video Transformer architectures can be classified based on the embeddings (backbone and minimal embeddings), tokenization (patch tokenization, frame tokenization, clip tokenization), and positional embeddings. We chose the Video Vision Transformer(ViViT) \cite{arnab2021vivit} for our experiments due to its representation of the 3D convolution in the form of Tubelet embedding. ViViT is easily reproducible and has a good balance between the parameters and accuracy for small datasets. Moreover, ViViT-H scores an accuracy of 95.8, just below the 95.9 accuracy score by Swin-L as per the Video Transformers Survey\cite{selva2022video} over HowTo100M\cite{miech2019howto100m}.
\begin{figure}[h]
\centering
\includegraphics[width=90mm]{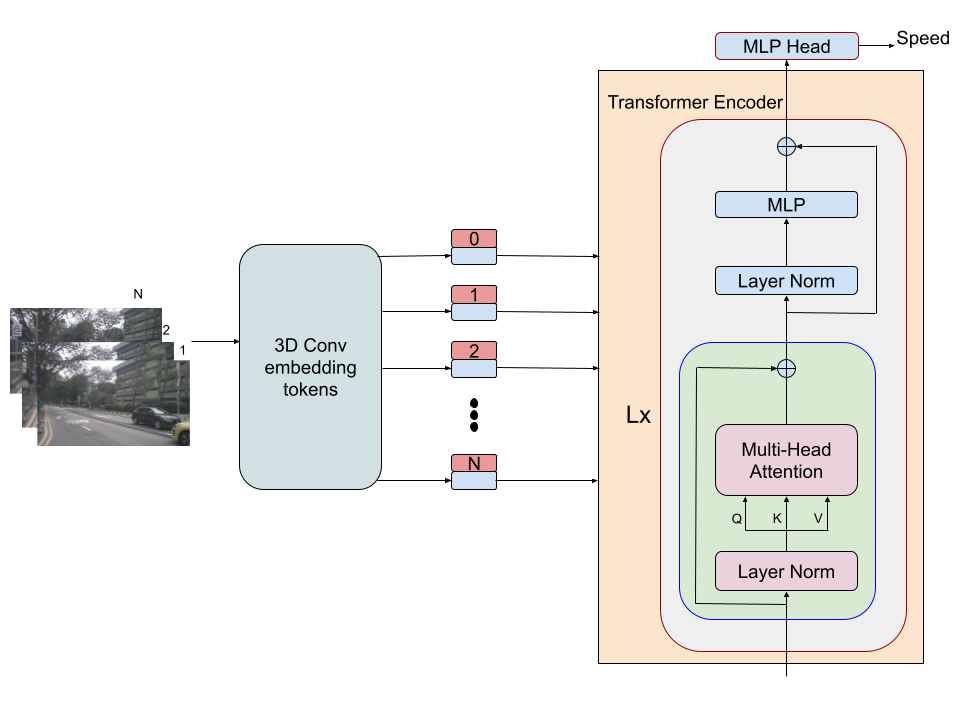} 
\caption{Architecture of ViViT. Here the frames from the video(N) are tokenized using 3D-Convolutional tubelet embeddings and further passed to multiple transformer encoders to regress the speed value finally. The Transformer Encoder is trained with the spatio-temporal embeddings}
\label{fig:vivit}
\end{figure}

\section{METHODS}
We aim to estimate the ego vehicle speed by relying purely on video streams from a monocular camera. A time-series model must be utilized to capture the relative motion between adjacent image data samples. The authors of \cite{https://doi.org/10.48550/arxiv.1412.0767} have proved the capability of a 3D-CNN to learn spatio-temporal features.

A 2D convolution operation over an image $I$ using a kernel $K$ of size m$\times$ n is given by \cite{GoodBengCour16} as :

\begin{align}
S(i,j) &= (I*K)(i,j) \\
&=\sum_{m}^{}\sum_{n}^{}I(i,j)K(i-m,j-n)
\end{align}

Expanding further on the above equation, the 3D convolution operation can be expressed as : 
\begin{align}
S(h,i,j) &= (I*K)(h,i,j) \\
&=\sum_{l}^{}\sum_{m}^{}\sum_{n}^{}I(h,i,j)K(h-l,i-m,j-n)
\end{align}

where $h$ is the additional dimension that includes the number of frames the kernel has to go through. Here the kernel is convoluted with the concatenation of the grayscale images and lane line segmentation masks.

To this extent, we incorporate a 3D-CNN network to preserve the temporal information of the input signals and compute the ego vehicle speed. 3D-CNNs can learn spatial and temporal features simultaneously using 3D kernels\cite{ji20123d}. We use small receptive fields of $3 \times 3 \times 3$ as our convolutional kernels throughout the network. Many 3D-CNN architectures lose big chunks of temporal information after the first 3D pooling layer. This is especially valid in the case of short-term spatio-temporal features propagated by utilizing smaller temporal windows. We refer to the pooling kernel size as $d \times k \times k$, where $d$ is the kernel temporal depth, and $s$ is the spatial kernel size. Similar to \cite{https://doi.org/10.48550/arxiv.1412.0767}, we used $d = 1$ for the first max pooling layer to preserve the temporal information. This way, we ensure that the temporal information does not collapse entirely after the initial convolutional layers.

In this paper, our contribution includes adding a masked-attention layer into the 3D-CNN architecture to guide the model to focus on relevant features that help with ego-vehicle speed computation. An image of an outdoor scene captured from a moving car has significant clutter and random motion that can obscure the model learning. We believe that the model can learn to filter out the irrelevant movements (such as that of other cars, pedestrians, etc.) that do not contribute towards the ego-vehicle speed estimation and focus only on features that matter if large quantities of data are available. However, in the more practical scenario where unlimited resources are not at our disposal, we think adding masked-attention helps to attain improved model performance with faster model convergence. We show that the error in speed estimation reduces by adding masked-attention to the 3D-CNN network. Further details about the impact of masked-attention are described as part of the ablation study in section~\ref{subsection:ablation}.

\begin{figure}[h]
\centering
\includegraphics[width=90mm]{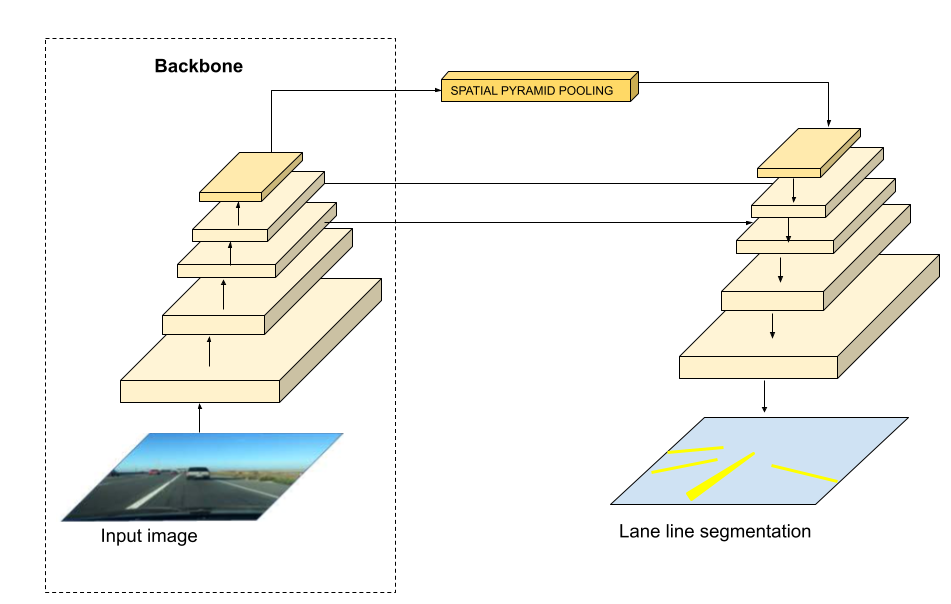} 
\caption{Architecture from \cite{2108.11250} modified for Lane line segmentation comprises of an encoder and a decoder }
\label{fig:yolop}
\end{figure}

\subsection{Masked-Attention}

Convolutional neural networks comprise a learned set of filters, where each filter extracts a different feature from the image\cite{https://doi.org/10.48550/arxiv.1603.07285}.
We aim to inhibit or exhibit the activation of features based on the appearance of objects of interest in the images. Typical scenes captured by car-mounted imaging devices include background objects such as the sky, and environment vehicles, which do not contribute to ego-vehicle speed estimation.  In fact, the relative motion of environmental vehicles often contributes negatively to the ability of the neural network to inhibit irrelevant features.

To inhibit and exhibit features based on relevance, we concatenate the masked-attention map to the input image before passing it through the neural network. We utilize a single-shot network with a shared encoder and three separate decoders that accomplish specific tasks such as object detection, drivable area segmentation, and lane line segmentation\cite{2108.11250}. There are no complex and redundant shared blocks between different decoders, which reduces computational consumption. CSP-Darknet\cite{https://doi.org/10.48550/arxiv.2011.08036} is chosen as the backbone network of the encoder, while the neck is mainly composed of Spatial Pyramid Pooling (SPP) module\cite{He_2014} and Feature Pyramid Network (FPN) module\cite{https://doi.org/10.48550/arxiv.1612.03144}. SPP generates and fuses features of different scales, and FPN fuses features at different semantic levels, making the generated features contain multiple scales and semantic level information.

The masked-attention map is generated from input video sequences using the lane line segmentation branch. The concatenation of lane segmentation as an additional channel to the camera channel allows the 3D-CNN to focus on the apparent displacement of the lane line segments in the video sequences to best estimate the ego-vehicle speed. Fig.~\ref{fig:yolop} shows the modified architecture designed by \cite{2108.11250} for extraction of lane line segments. 

\begin{figure*}[h]
\begin{center}
\includegraphics[width=165mm]{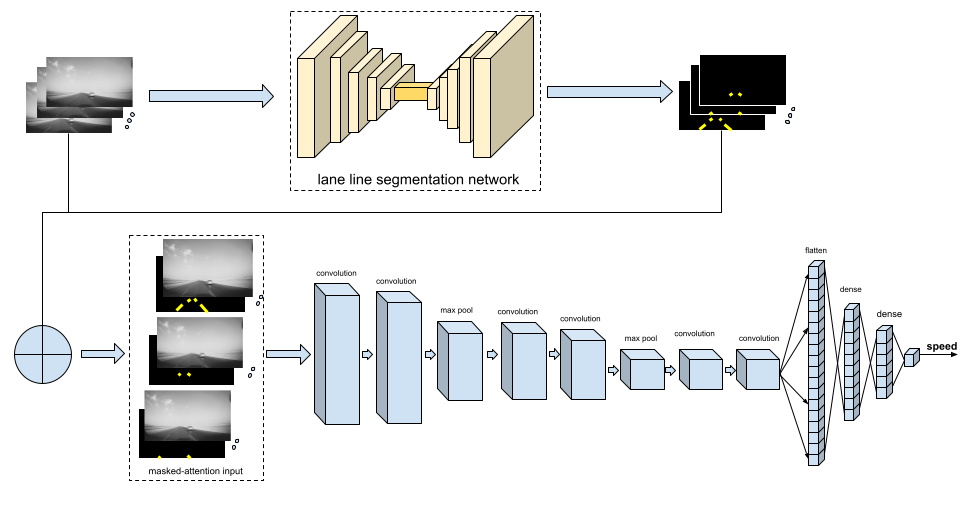} 
\caption{architecture of 3DCMA}
\label{fig:networkarch}
\end{center}
\end{figure*}
\subsection{Network Architecture}
Our 3D-CNN architecture with masked-attention for ego vehicle speed estimation is illustrated in Fig.~\ref{fig:networkarch}.

We convert the RGB stream to grayscale since color information is not vital for speed estimation.  However, masked-attention map is concatenated as an additional channel to the grayscale image. To reduce the computational complexity and memory requirement, the original input streams are resized to $64 \times 64$ before feeding them into the network. Thus the input to the model has a dimension of $n \times 64 \times 64 \times 2$, where $n$ is the number of frames in the temporal sequence. All convolutional 3D layers use a fixed kernel size of $3 \times 3 \times 3$ as recommended in \cite{https://doi.org/10.48550/arxiv.1412.0767}. The initial pooling layer uses a kernel size of $1 \times 2 \times 2$ to preserve the temporal information. The subsequent pooling layer, which appears at the center of the network, compresses the temporal and spatial domains with a kernel size of $2 \times 2 \times 2$. 
We incorporate six 3D convolutional layers with the number of filters for the layers from $1-6$ being $ 32, 32, 64, 64, 128, 128$ respectively. Finally, four fully connected layers have $512, 256, 64$ and $1$ nodes.
% A summary of the model architecture is shown in Table~\ref{table:architecture}.

The L2 loss function which we used for the 3D-CNN can be described as :

\begin{align}
\mathcal{L}_{speed} &= \frac{1}{n}\sum_{i = 0}^{n}(S_{i}-\hat{S}_{i})^{2} \\
& = \frac{1}{n}\sum_{i = 0}^{n}(S_{i}-W^{T}X)^{2} \\
& = \frac{1}{n}\sum_{i = 0}^{n}(S_{i}-W^{T}(X_{I}+X_{M}))^{2}
\end{align}

where $n$ is the number of frames in the input and $S_i$ is the speed value ground truth of $i$th corresponding frame, and $\hat{S}_{i}$ is the inferred speed value. $X_I$ is the grayscale image channel, and $X_M$ is the masked-attention channel for every frame. $W$ is the weight tensor of the 3D convolutional kernel.

\section{EXPERIMENTATION}
Several experiments have been conducted to evaluate the effectiveness of our proposed AI model. We start by describing the public datasets used in our experiments. We then touch upon the metrics used for our evaluation. We compare our model architecture against ViViT which is amongst the state-of-the-art in the recent vision transformer architectures. We additionally conduct some ablation studies to understand the contribution of proposed masked-attention within our network architecture and compare its performance by discarding the same from 3D-CNN.

\begin{figure*}[h]
    \begin{subfigure}{1.0\textwidth}
        \centering
        \includegraphics[width=.19\textwidth]{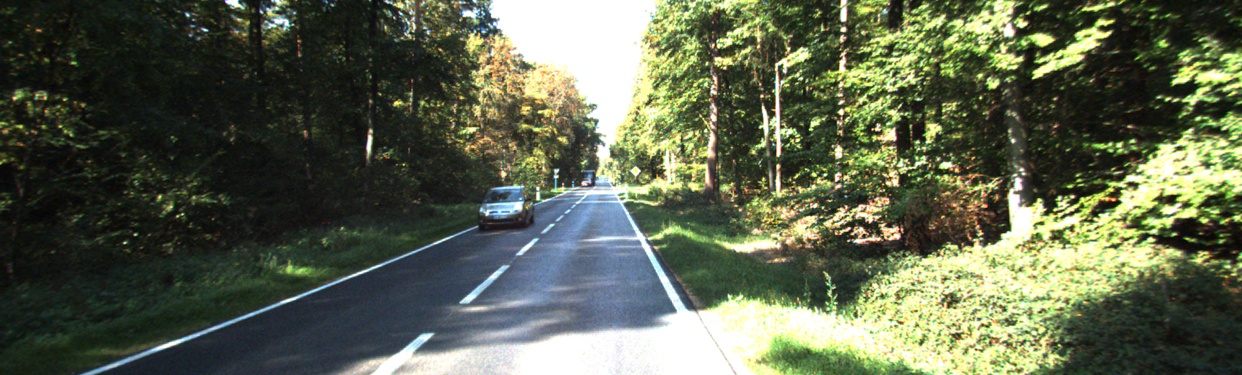}
        \includegraphics[width=.19\textwidth]{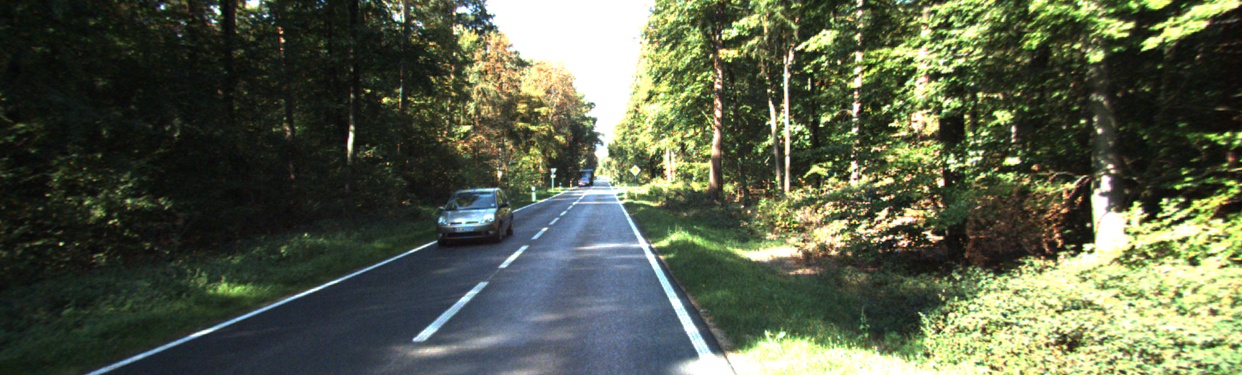}
        \includegraphics[width=.19\textwidth]{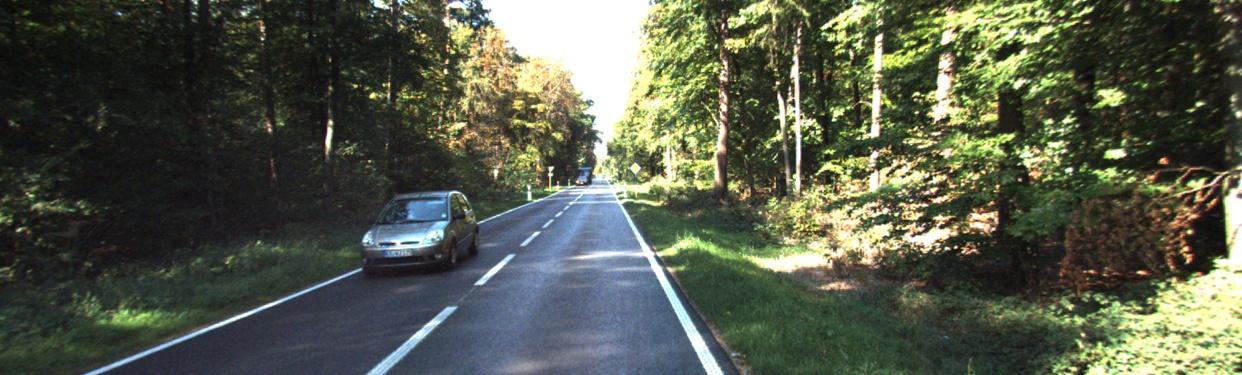}
        \includegraphics[width=.19\textwidth]{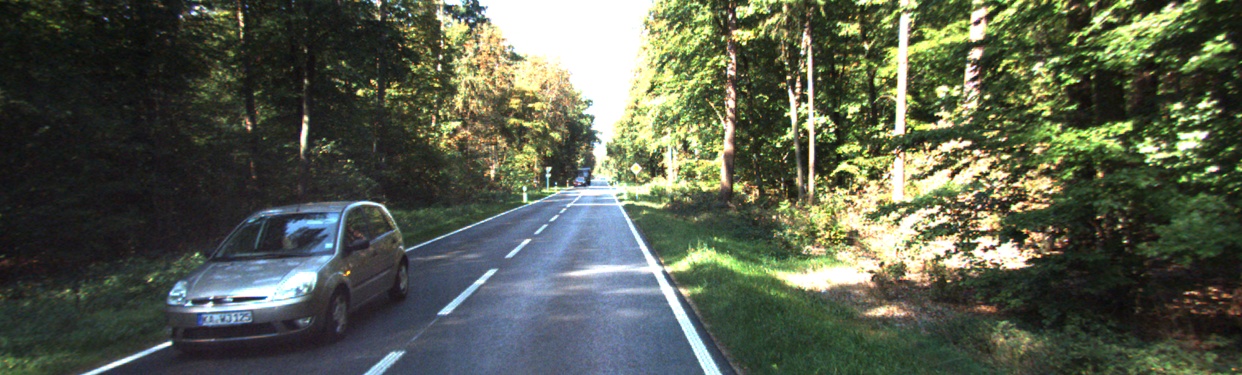}
        \includegraphics[width=.19\textwidth]{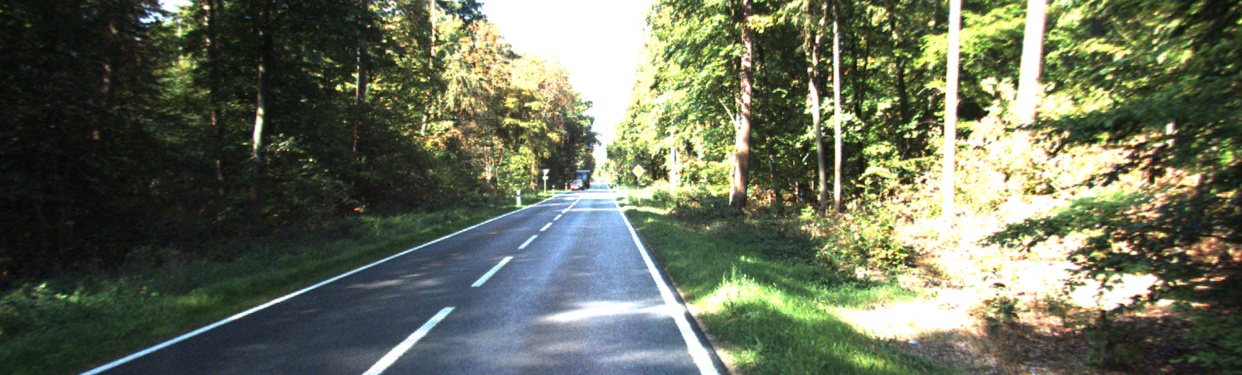}
        \caption{Road category images from KITTI dataset}
    \end{subfigure}
    \hfill
    \begin{subfigure}{1.0\textwidth}
        \centering
        \includegraphics[width=.19\textwidth]{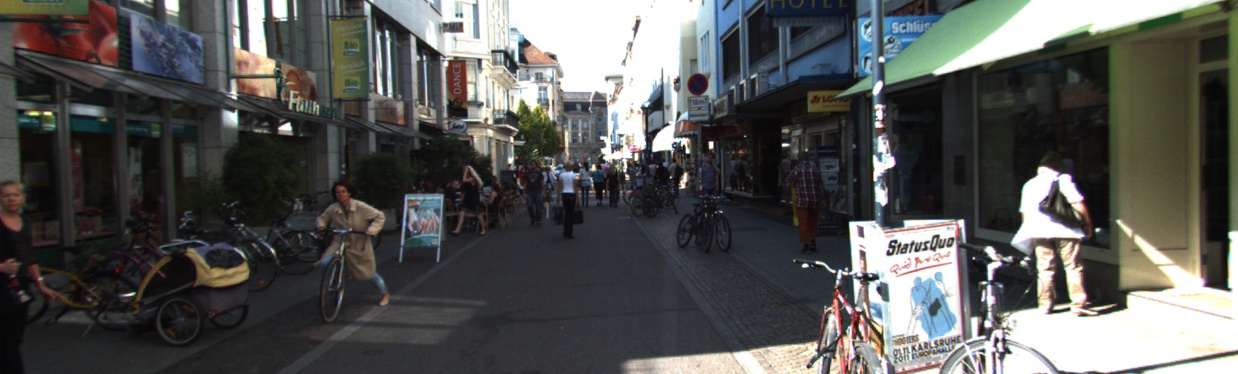}
        \includegraphics[width=.19\textwidth]{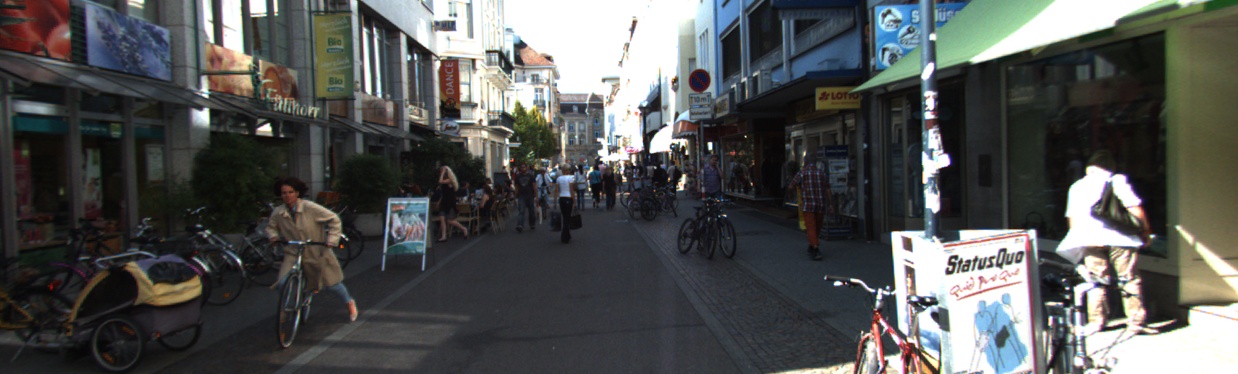}
        \includegraphics[width=.19\textwidth]{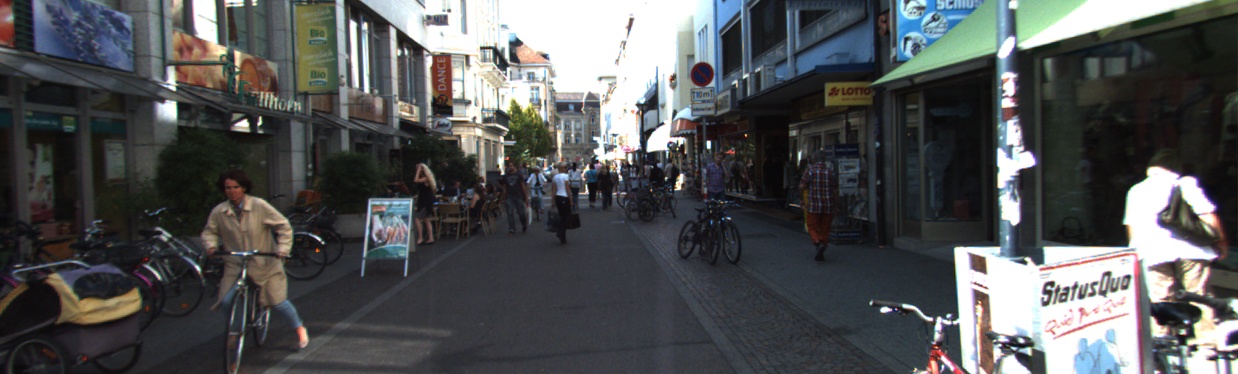}
        \includegraphics[width=.19\textwidth]{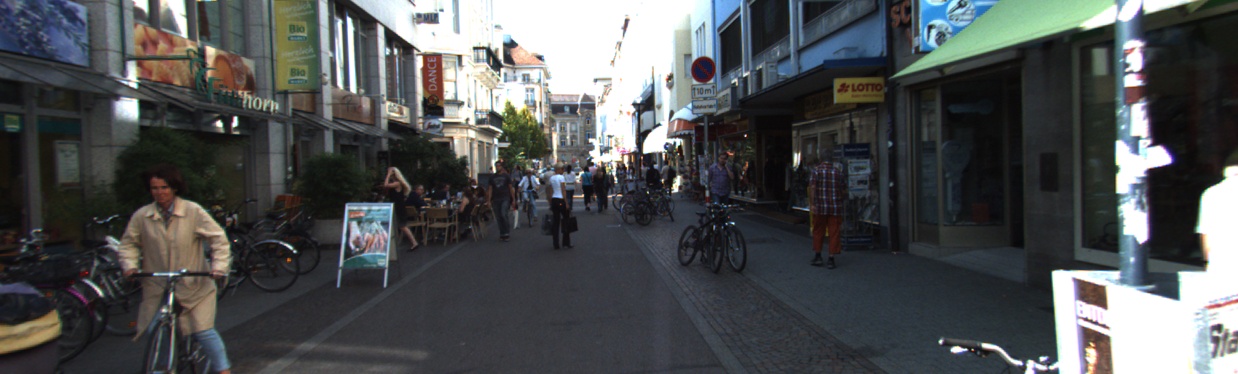}
        \includegraphics[width=.19\textwidth]{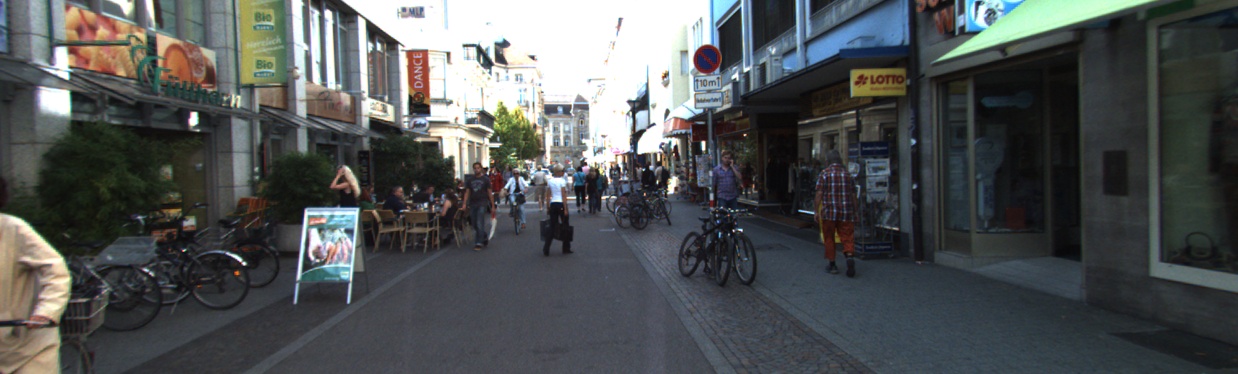}
        \caption{City category images from KITTI dataset}
    \end{subfigure}
    \hfill
    \begin{subfigure}{1.0\textwidth}
        \centering
        \includegraphics[width=.19\textwidth]{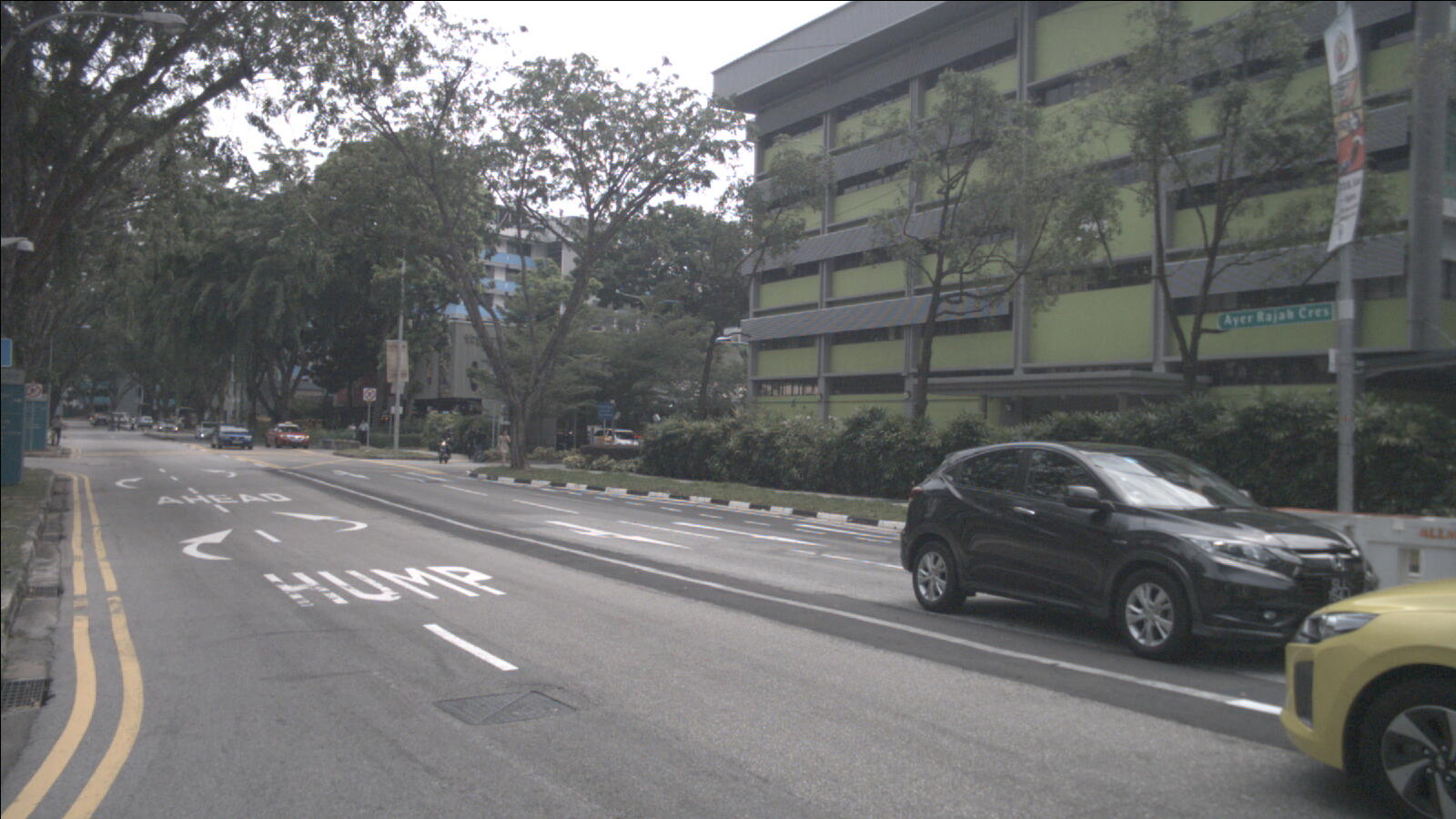}
        \includegraphics[width=.19\textwidth]{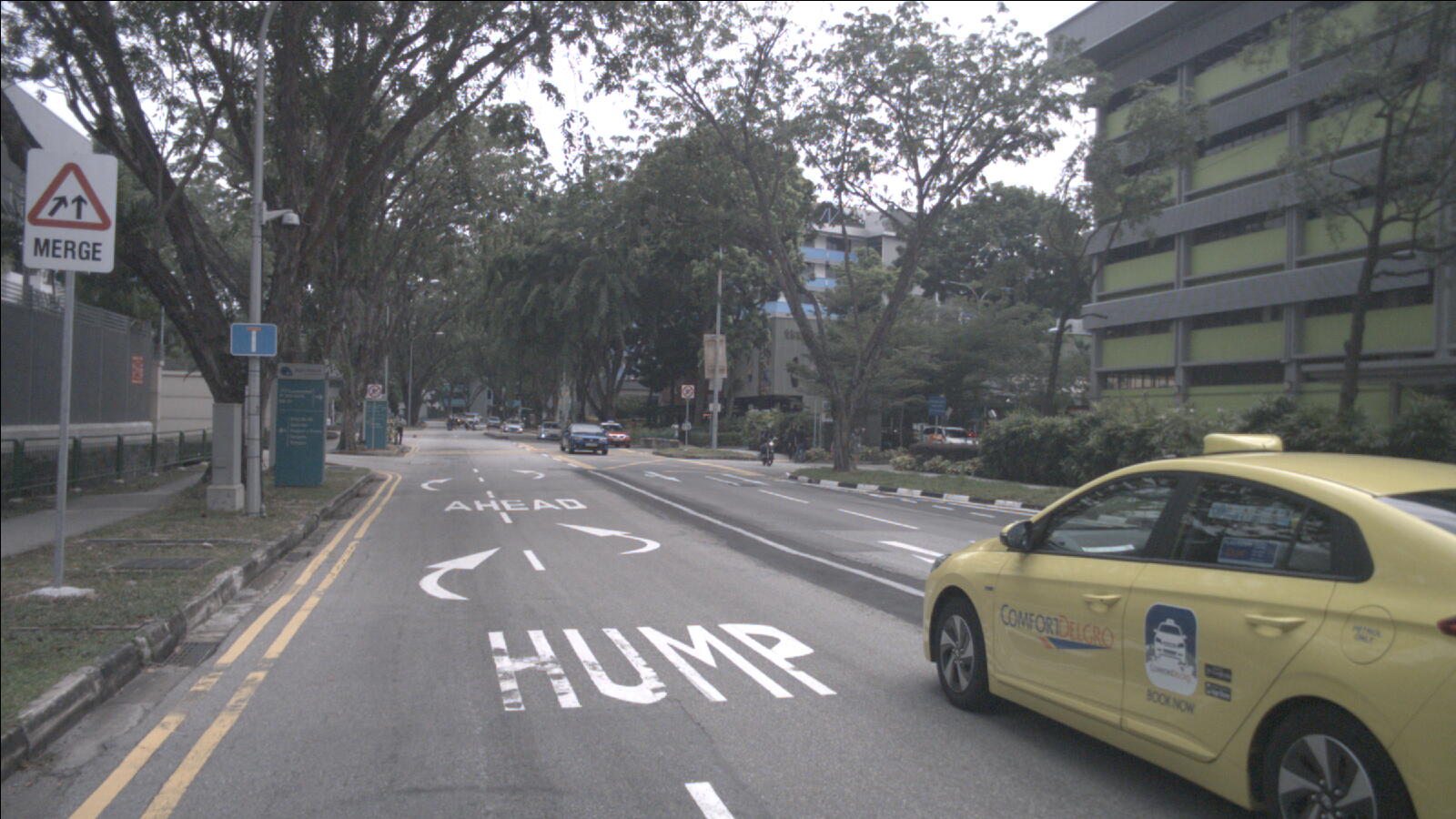}
        \includegraphics[width=.19\textwidth]{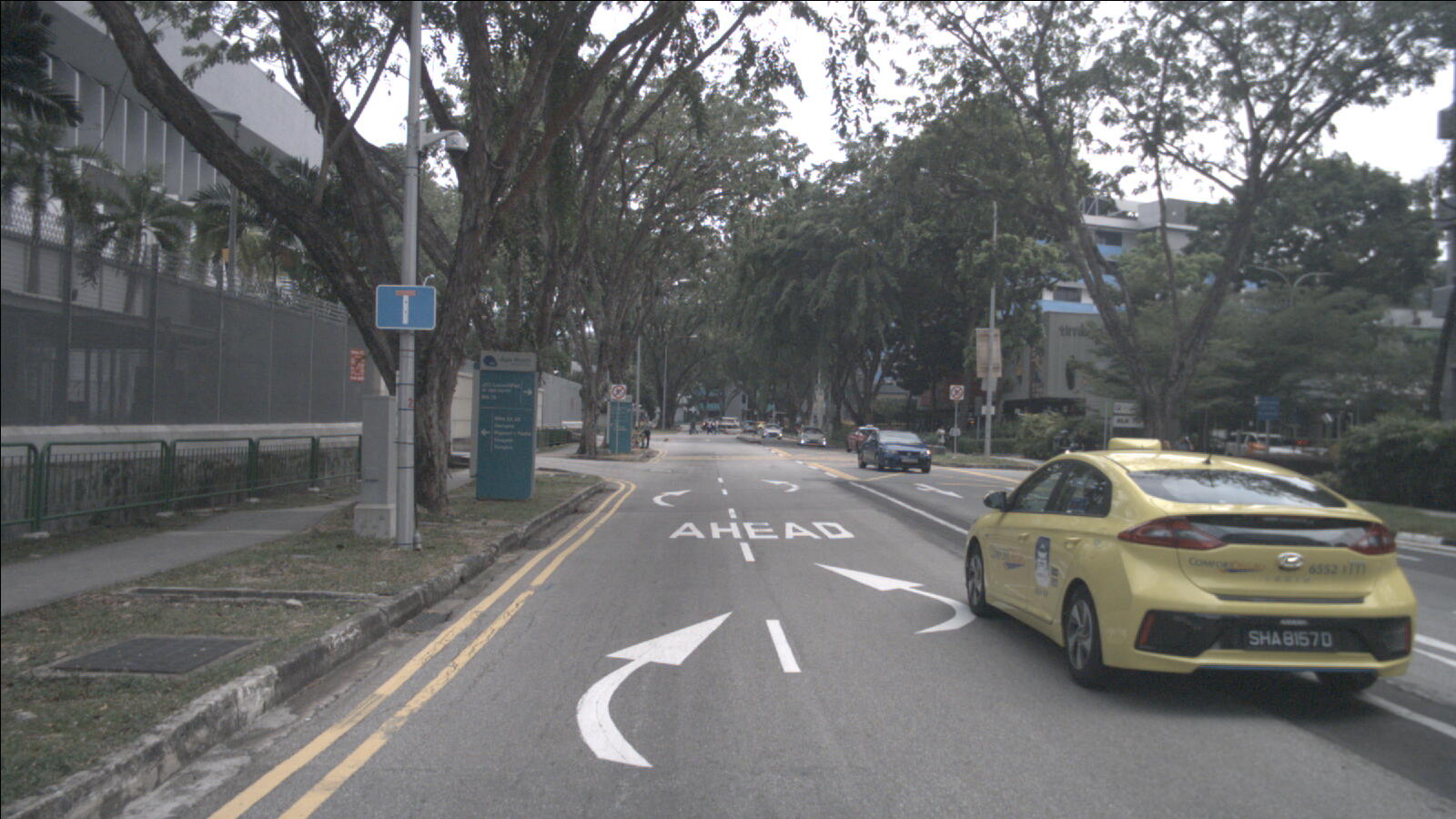}
        \includegraphics[width=.19\textwidth]{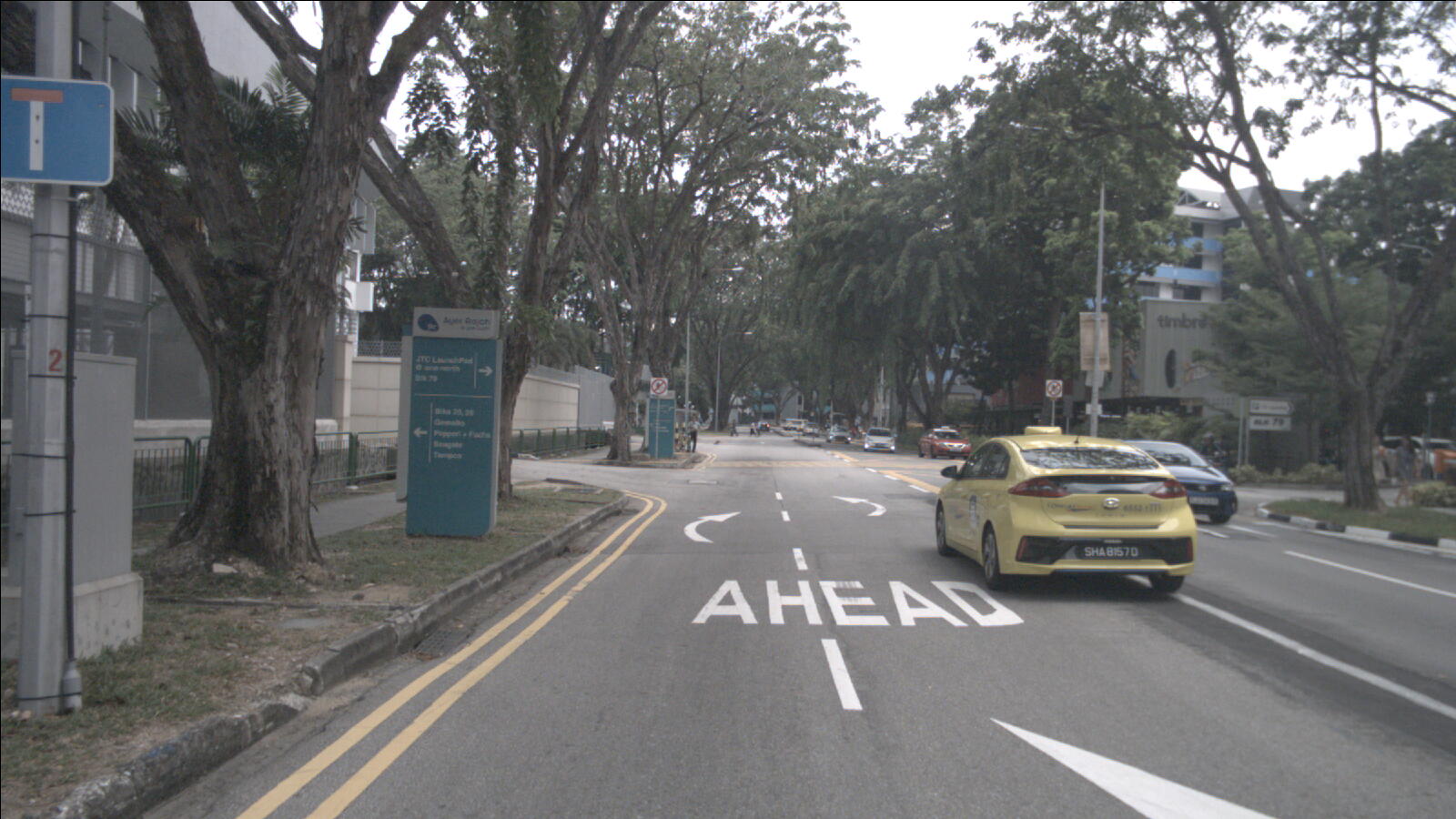}
        \includegraphics[width=.19\textwidth]{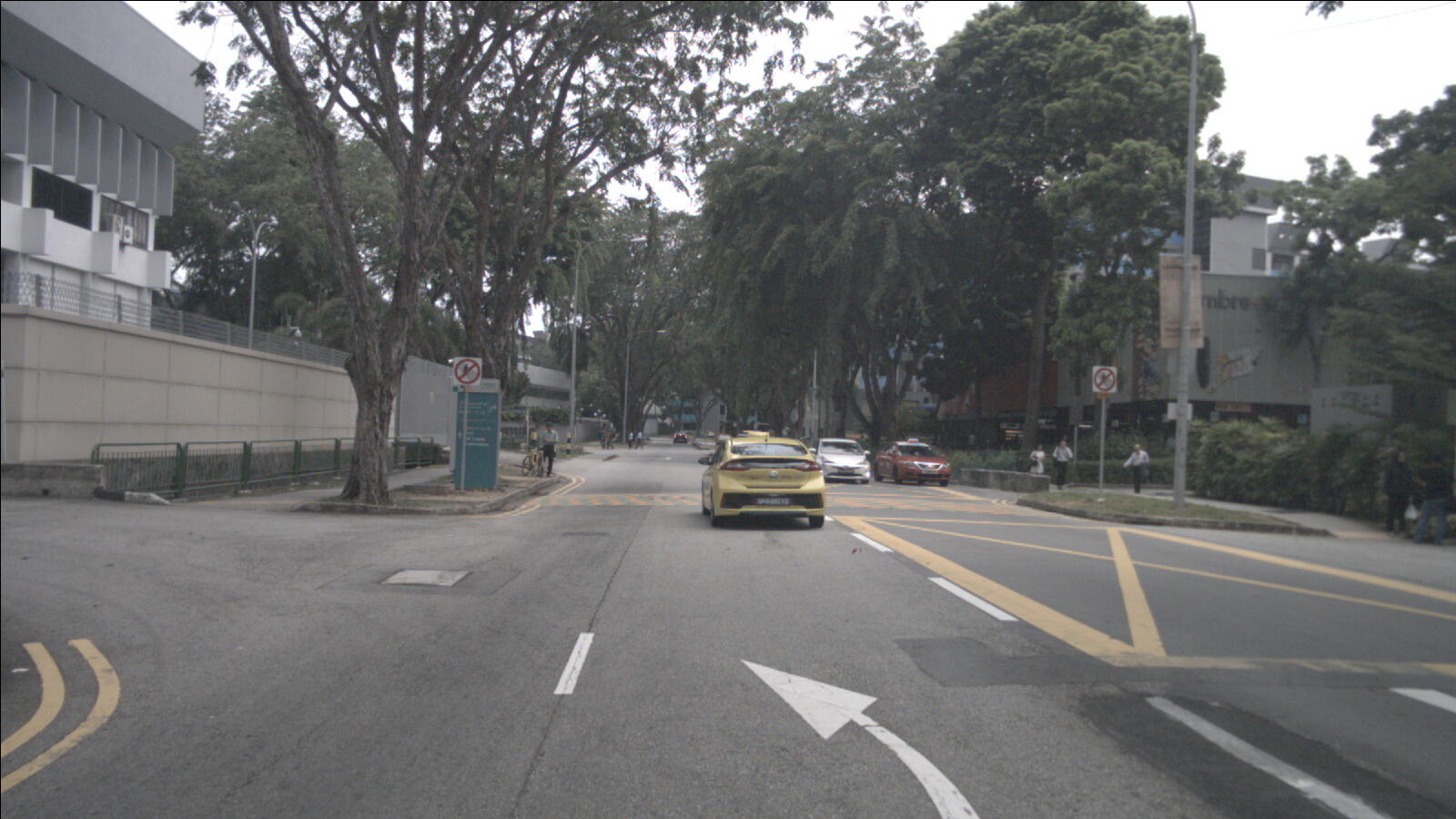}
        \caption{Daytime images from nuImages dataset}
    \end{subfigure}
    \hfill
    \begin{subfigure}{1.0\textwidth}
        \centering
        \includegraphics[width=.19\textwidth]{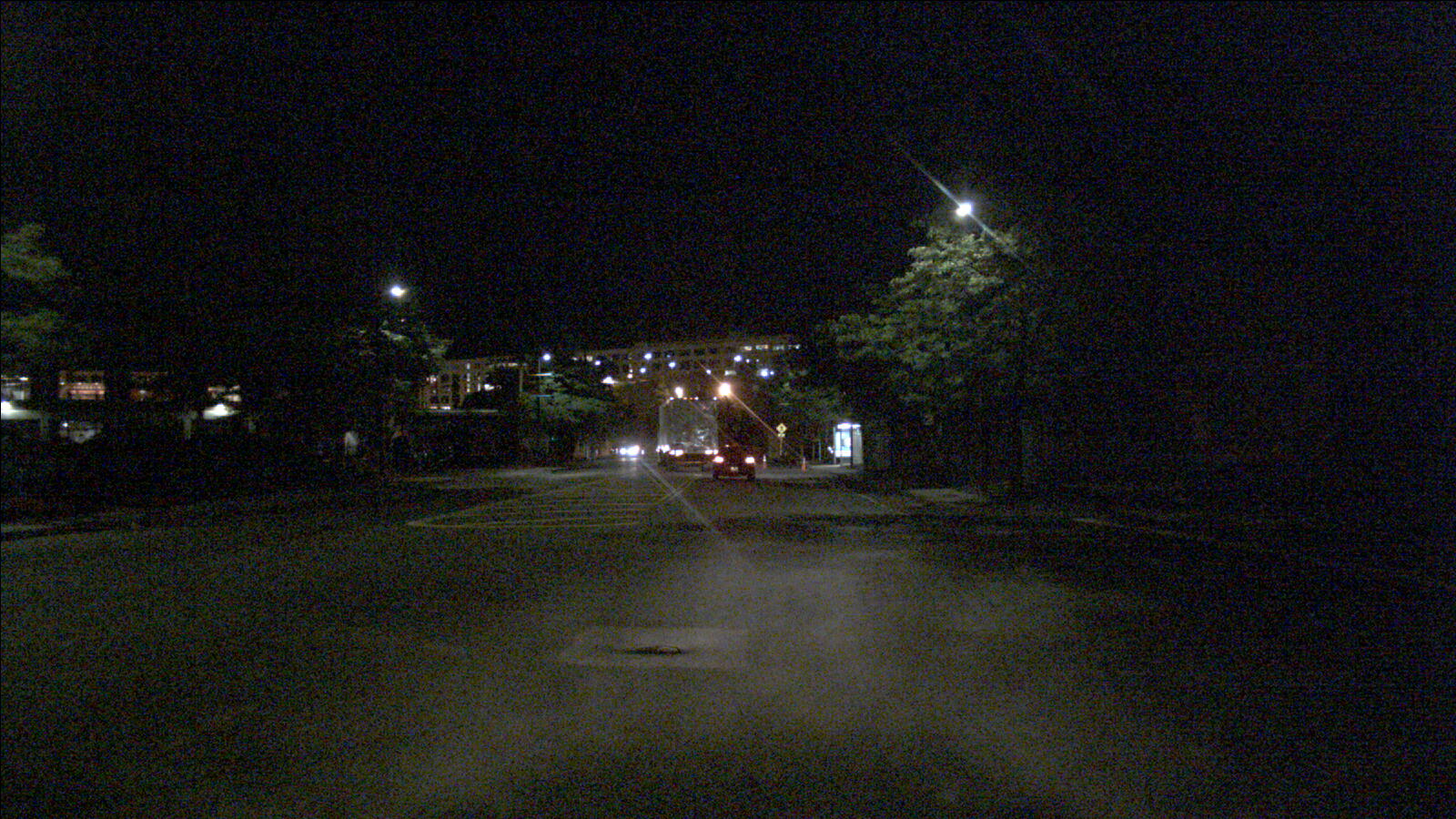}
        \includegraphics[width=.19\textwidth]{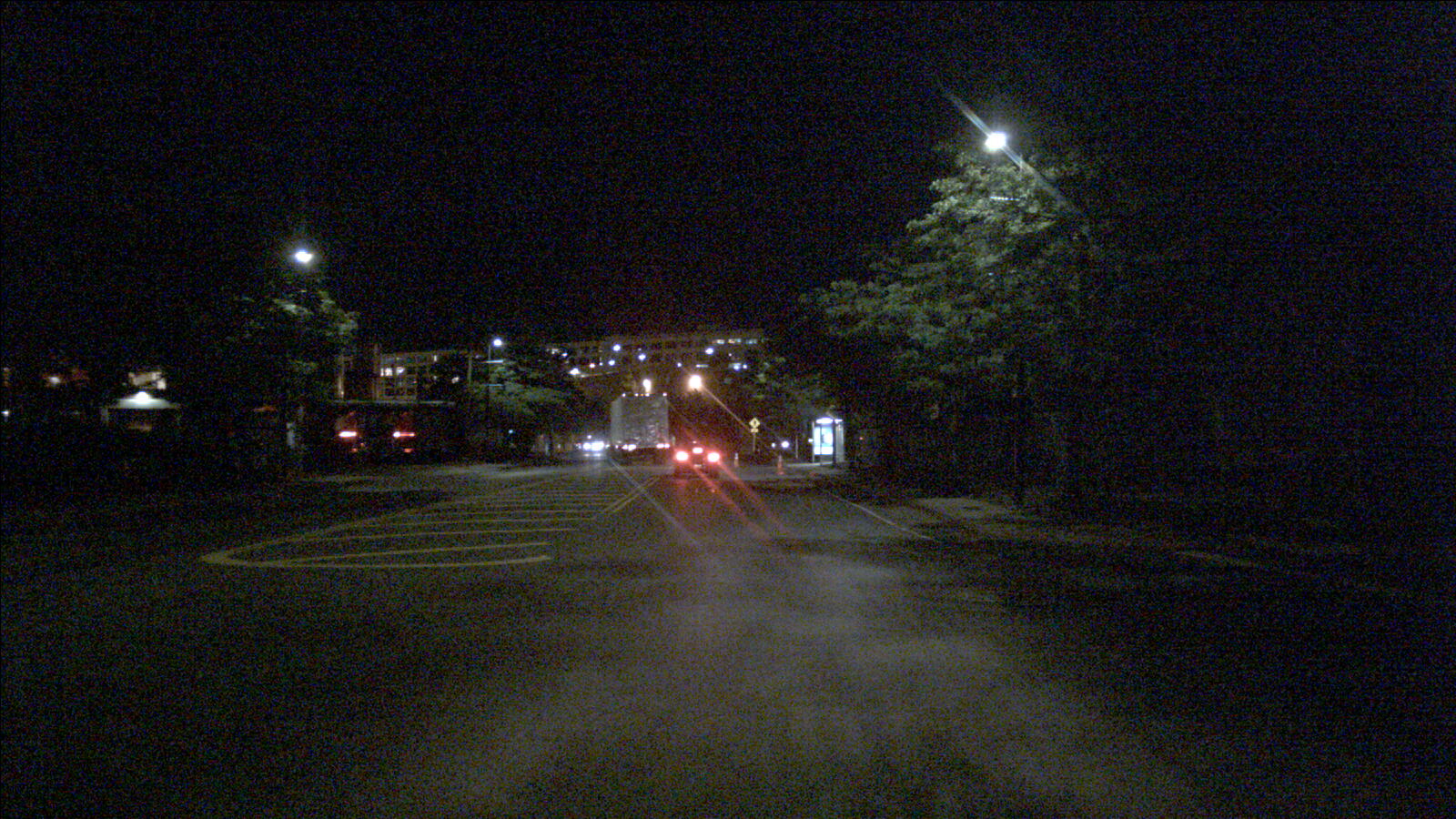}
        \includegraphics[width=.19\textwidth]{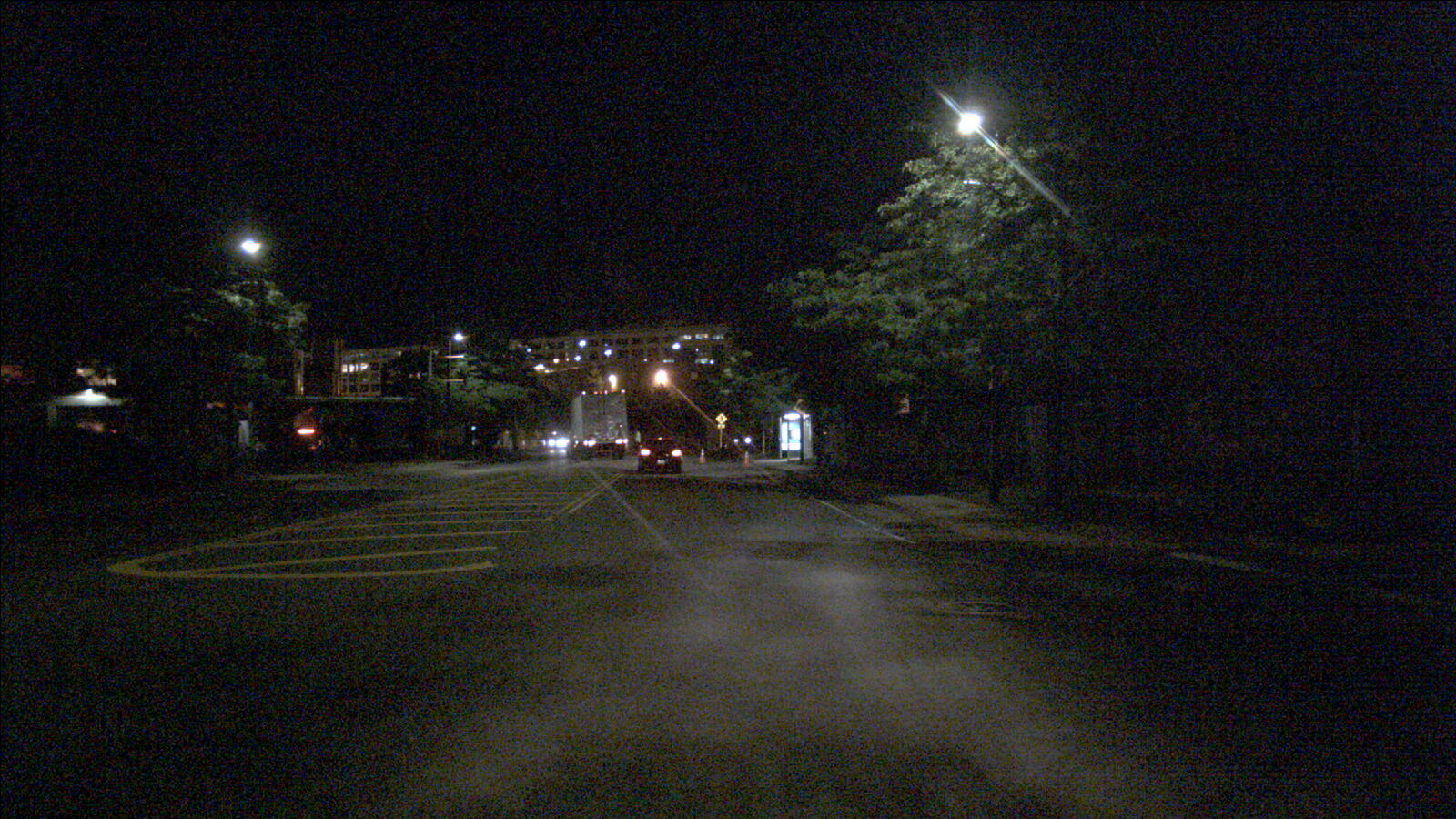}
        \includegraphics[width=.19\textwidth]{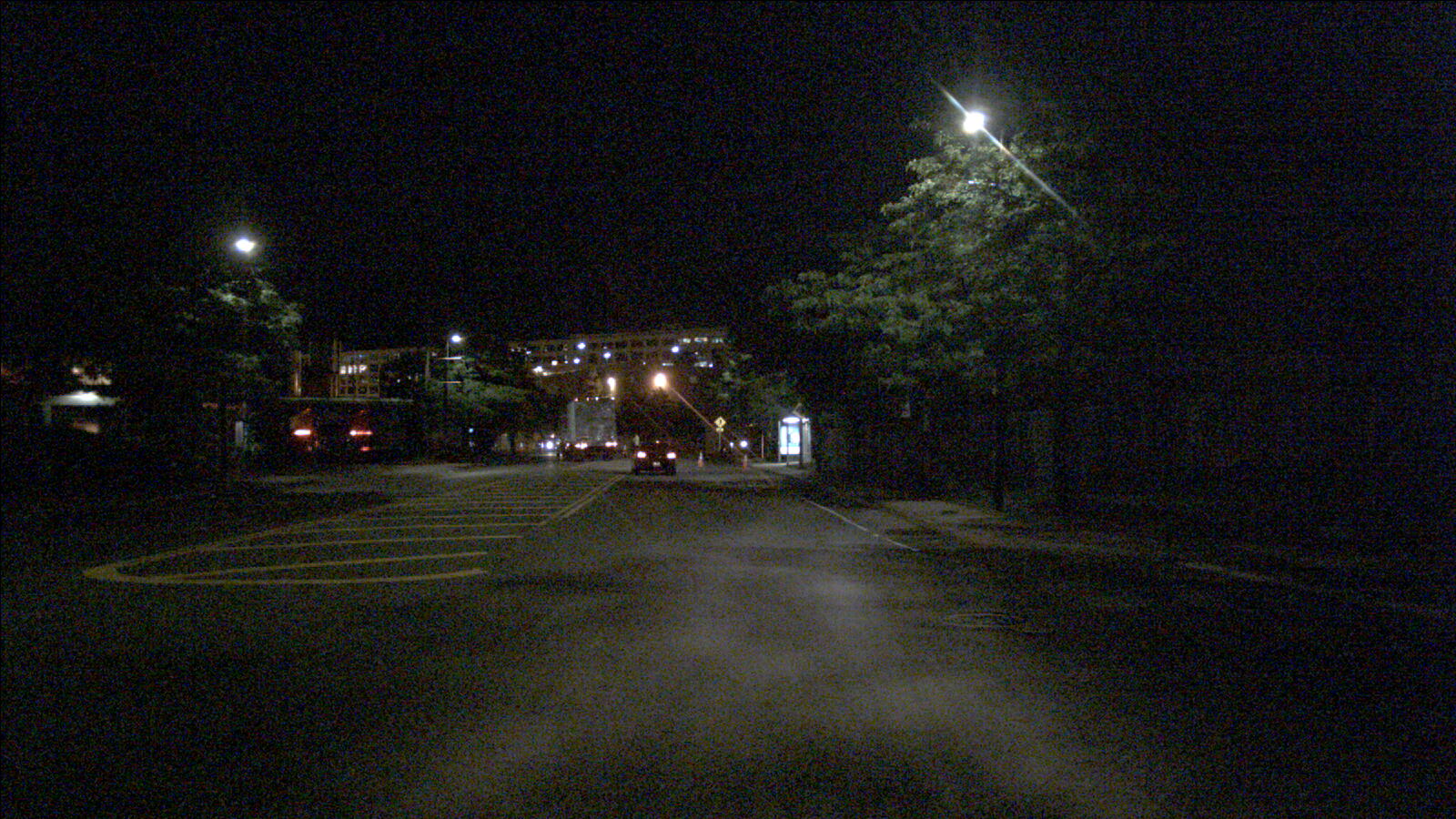}
        \includegraphics[width=.19\textwidth]{images/nuimages_night/11.jpg}
        \caption{Night time images from nuImages dataset}
    \end{subfigure}

\caption{Visualization of sample images - KITTI and nuImages dataset}
\label{pics:sampleimages}
\end{figure*}

\subsection{Datasets}
As per our knowledge, there is a lack of standardized datasets available in the literature for the estimation of ego-vehicle speed from a monocular camera stream. DBNet\cite{8578713} is a large-scale dataset for driving behavior research which includes aligned videos and vehicular speed from 1000 km driving stretch. However, the test set is not available for public usage to the best of our knowledge. Likewise, the test set of comma.ai speed challenge\cite{commaspeedchallenge} is not open to the public at the time of writing this paper. The authors in \cite{https://doi.org/10.48550/arxiv.1907.06989} utilized KITTI dataset for speed estimation using motion and monocular depth estimation. However, there is no information about the train and test splits used for the evaluation of the models. In this paper, we decided to utilize two public datasets for our experiments - nuImages and KITTI. Some sample images extracted from video sequences for nuImages and KITTI are shown in Fig.~\ref{pics:sampleimages}.

\begin{figure*}[h]
    \centering
    \includegraphics[width=1.0\textwidth]{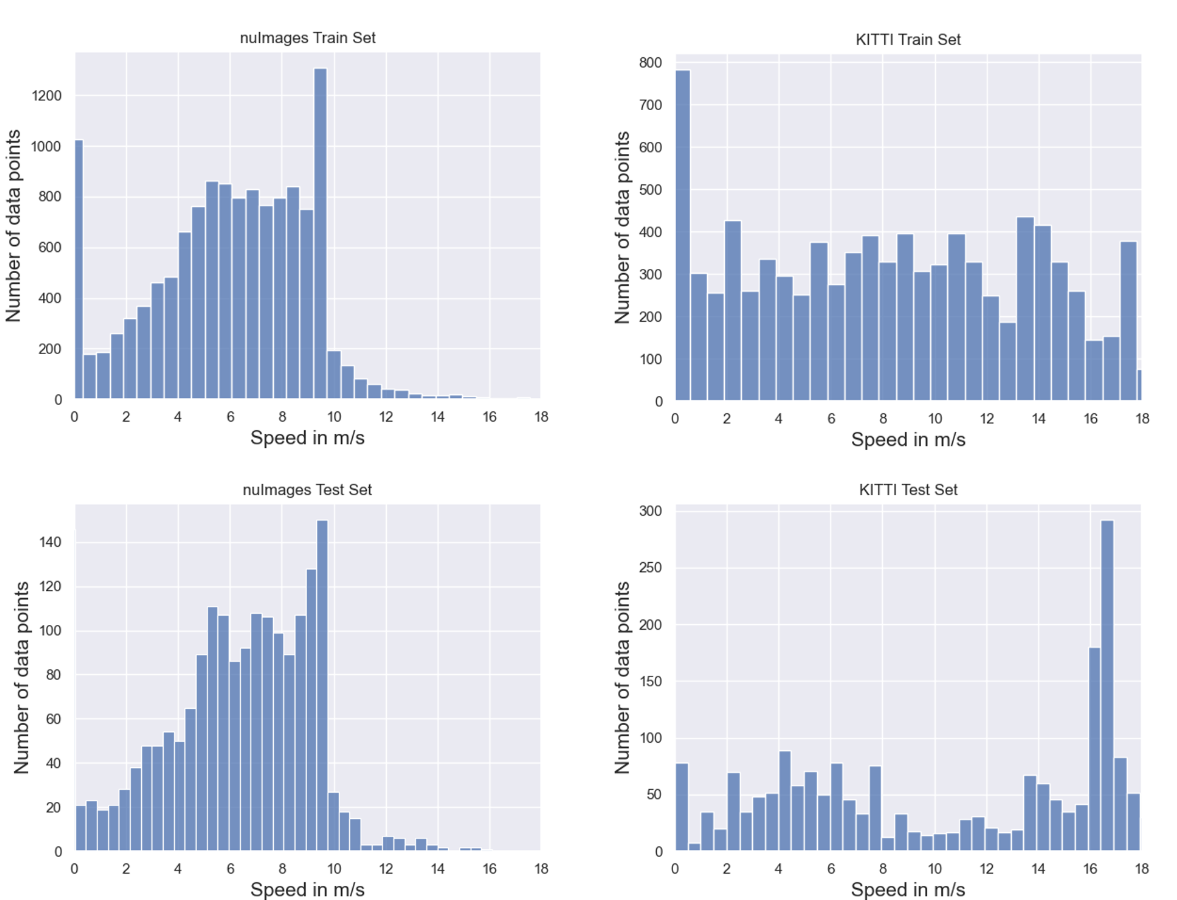}
    \caption{train/test speed data distribution for nuImages and KITTI datasets}
    \label{pics:datasplit}
\end{figure*}

\subsubsection{nuImages Dataset}
nuImages is derived from nuScenes \cite{nuscenes}, and it is a large-scale autonomous driving dataset having 93k video clips of 6 seconds each. The dataset is collated from two diverse cities - Boston and Singapore. Each video clip consists of 13 frames spaced out at 2 Hz. The annotated images include rain, snow, and night time, which are important for autonomous driving applications.

Each sample in the nuImages dataset comprises of an annotated camera image with an associated timestamp and past and future images. It is to be noted that the six previous and six future images are not annotated. The sample frame has meta-data information available as token ids regarding the previous and future frames associated with the particular sample. 

% \begin{figure}[h]
%     \centering
%     \subfloat{{\includegraphics[width=8cm, height = 4cm]{./images/nuimages_trainset.png} }}%
%     \qquad
%     \subfloat{{\includegraphics[width=8cm, height = 4cm]{./images/nuimages_testset.png} }}%
%     \caption{train/test speed data distribution for nuImages dataset}%
%     \label{fig:nuimagesdata}%
% \end{figure}

The vehicle speed is extracted from the CAN bus data and linked to the sample data through sample tokens. We have strictly followed the train and test splits of the nuImages dataset for training and evaluating the AI models. The distribution of speed data across train and test splits of the nuImages dataset are shown in Fig.~\ref{pics:datasplit}.

\subsubsection{KITTI Dataset}

The KITTI Vision Benchmark Suite \cite{6248074,doi:10.1177/0278364913491297} is a public dataset containing raw data recordings that are captured and synchronized at 10 Hz. \cite{6248074} presented the benchmark challenges, their creation and use for evaluating state-of-the-art computer vision methods, while \cite{doi:10.1177/0278364913491297} was a follow-up work that provided technical details on the raw data itself, describing the recording platform, the data format and the utilities.

The dataset was captured by driving around the mid-size city of Karlsruhe. We utilized the "synched+rectified" processed data where images are rectified and undistorted and where the data frame numbers correspond across all sensor streams. While the dataset provides both grayscale and color stereo sequences, we have resorted to utilizing an RGB stream extracted from camera ID $03$ only. The ego-vehicle speed values are extracted from IMU sensor readings. The raw data is split across six categories - City, Residential, Road, Campus, Person, and Calibration. For our experiment, we utilize data from City and Road categories. We noticed that some video samples in the City category have prolonged periods where the car is stationary. We thus discarded such video samples where the vehicle was stationary for most of the video samples. To facilitate future benchmarks from the research community for ego-vehicle speed estimation, we also report our train and test splits in Table~\ref{table:kitidata}. The distribution of speed data across train and test splits from the KITTI dataset is shown in Fig.~\ref{pics:datasplit}.

\begin{table}[ht]
\small
\begin{center}
\begin{tabular}{| *{4}{c|} }
\hline
KITTI  & \multicolumn{2}{c|}{}  &  \\
Category & \multicolumn{2}{c|}{Train} & Test \\
\hline
\multirow{8}{2em}{City} & \multirow{6}{8em}{2011\_09\_26\_drive\_} & 0002, 0005, 0009 & \\ 
& & 0011, 0013, 0014 & \\ 
& & 0048, 0051, 0056 &0001  \\ 
& & 0059, 0084, 0091 &0117  \\ 
& & 0095, 0096, 0104 &  \\ 
& & 0106, 0113 &  \\  
\cline{2-4}
&  \multirow{1}{8em}{2011\_09\_28\_drive\_} & 0001 &  \\
\cline{2-3}
& \multirow{1}{8em}{2011\_09\_29\_drive\_} & 0071 & \\
\hline
\multirow{4}{2em}{Road} & \multirow{2}{8em}{2011\_09\_26\_drive\_} & 0015, 0027, 0028  &  0070\\ 
& & 0029, 0032, 0052 &  0101 \\ 
\cline{2-4}
& \multirow{2}{8em}{2011\_09\_29\_drive\_} & 0004, 0016, 0042 &  \\ 
&  & 0047 & \\ 

\hline

\end{tabular}
\captionof{table}{train and test video samples for KITTI dataset} \label{table:kitidata}
\end{center}
\end{table}

\subsection{Evaluation metrics}
We utilize the conventional evaluation protocol that is used in the literature for the task of regression - Mean Absolute Error (MAE) and Root Mean Square Error (RMSE) [46]. 
\\

We compute the MAE and RMSE as follows :

\begin{equation}
RMSE = \sqrt{(\frac{1}{n})\sum_{i=1}^{n}(y_{i} - \hat{y}_{i})^{2}}
\end{equation}

\begin{equation}
MAE = (\frac{1}{n})\sum_{i=1}^{n}|\hat{y}_i-y_i|
\end{equation}

where $y_i$ denotes the ground truth ego-vehicle speed value and $\hat{y}_i$ denotes the predicted speed value by the AI model.

\subsection{Experiments}
\label{section:experiments}
We used RGB images from the camera mounted in front of the vehicle and ego-vehicle velocity coming from the CAN-BUS across both public datasets. This information is synchronized. The KITTI dataset has a camera image resolution of $1238 \times 374$. The temporal dimension we used for the KITTI dataset is ten frames. The KITTI dataset is sampled at 10 Hz, which means that the models are fed with video frames containing visual information from a time window of 1 [sec]. The ego-vehicle velocity assigned to any temporal sequence is the speed value tagged to the closest time stamp of the 10th frame in the input sequence. 

On the other hand, the camera image resolution for the nuImages dataset is $1600 \times 900$. nuImages dataset is sampled at 2 Hz. We take six frames each, preceding and succeeding the sample frame. This means that the models are fed with video frames containing visual information spanning a time window of approximately 6 [sec]. The ego vehicle velocity assigned to any temporal sequence is the speed value tagged to the closest time-stamp of the sample frame (7th frame in the input sequence)

For our experiments with ViViT, we extract non-overlapping, spatio-temporal tubelet embeddings of dimension $t \times h \times w$, where $ t = 6$, $h = 8$, and $w = 8$. The number of transformer layers in our implementation is $16$. The number of heads for multi-headed self-attention blocks is $16$, and the dimension of embeddings is $128$. 

\subsection{Setup}

Our AI models were trained using an Nvidia GeForce RTX-$3070$ Max-Q Design GPU having $8$ GB VRAM. The learning rate used for training all models is $1\times10^{-3}$. All models are trained for $100$ epochs with early stopping criteria set to terminate the training process if validation loss does not improve for ten epochs consecutively. The optimizer utilized is Adam since it utilises both momentum and scaling.

\section{RESULTS}
We evaluate the performance of our proposed 3DCMA architecture and compare it against the standard ViViT with spatio-temporal attention. We report the evaluation metrics on the test set for KITTI and nuImages datasets in the subsections below. Our evaluation across all datasets consistently reported better results for our proposed 3DCMA architecture.

\subsubsection{nuImages dataset}
Evaluation scores for the nuImages dataset are shown in Table~\ref{tab:table3}. We observed approximately $27\%$ improvement in RMSE and MAE for 3DCMA compared to ViViT for the nuImages dataset.

\begin{table}[h]
  \centering
  \begin{tabular}{l r r}
    % \toprule
    & \multicolumn{2}{c}{\small{\textbf{Evaluation Metric}}} \\
    \cmidrule(r){2-3}
    {\small\textit{Method}}
    & {\small \textit{RMSE}}
      & {\small \textit{MAE}}\\
    \midrule
    ViViT & 1.782 & 1.326\\
    \textbf{3DCMA} & \textbf{1.297} & \textbf{0.974}\\
    % \bottomrule
  \end{tabular}
  \caption{nuImages evaluation for (a)ViViT (b)3DCMA}~\label{tab:table3}
\end{table}

\subsubsection{KITTI dataset}

Our evaluation shows $34.5\%$ and $41.5\%$ improvement in RMSE and MAE respectively on the KITTI dataset for 3DCMA compared to ViViT. The results are seen in Table~\ref{tab:table4}.

\begin{table}[h!]
  \centering
  \begin{tabular}{l r r}
    % \toprule
    & \multicolumn{2}{c}{\small{\textbf{Evaluation Metric}}} \\
    \cmidrule(r){2-3}
    {\small\textit{Method}}
    & {\small \textit{RMSE}}
      & {\small \textit{MAE}}\\
    \midrule
    ViViT & 5.024 & 4.324\\
    \textbf{3DCMA} & \textbf{3.290} & \textbf{2.528}\\
    % \bottomrule
  \end{tabular}
  \caption{Evaluation on KITTI dataset for (a)ViViT (b)3DCMA}~\label{tab:table4}
\end{table}

\subsection{Ablation Study}
\label{subsection:ablation}

To further understand the importance of masked-attention, we conducted an ablation study by removing masked-attention input to the 3D-CNN network. It is to be noted that the input to the 3D-CNN model is a single-channel grayscale image after the removal of the masked-attention input. 

\subsubsection{nuImages dataset}
Evaluation scores for the nuImages dataset are shown in Table~\ref{tab:table5}. The addition of masked-attention reduces RMSE by $23.6\%$ and MAE by $25.9\%$ for the nuImages dataset.

\begin{table}[h!]
  \centering
  \begin{tabular}{l r r}
    % \toprule
    & \multicolumn{2}{c}{\small{\textbf{Evaluation Metric}}} \\
    \cmidrule(r){2-3}
    {\small\textit{Method}}
    & {\small \textit{RMSE}}
      & {\small \textit{MAE}}\\
    \midrule
    3D-CNN without MA & 1.698 & 1.315\\
    \textbf{3DCMA} & \textbf{1.297} & \textbf{0.974}\\
    % \bottomrule
  \end{tabular}
  \caption{Evaluation on nuImages dataset for (a)3D-CNN without masked-attention (b)3DCMA}~\label{tab:table5}
\end{table}

\subsubsection{KITTI dataset}
Evaluation scores for the KITTI dataset are shown in Table~\ref{tab:table6}. The addition of masked-attention reduces the RMSE by $25.8\%$ and MAE by $30.1\%$ for the KITTI dataset.

\begin{table}[h]
  \centering
  \begin{tabular}{l r r}
    % \toprule
    & \multicolumn{2}{c}{\small{\textbf{Evaluation Metric}}} \\
    \cmidrule(r){2-3}
    {\small\textit{Method}}
    & {\small \textit{RMSE}}
      & {\small \textit{MAE}}\\
    \midrule
    3D-CNN without MA & 4.437 & 3.617\\
    \textbf{3DCMA} & \textbf{3.290} & \textbf{2.528}\\
    % \bottomrule
  \end{tabular}
  \caption{Evaluation on KITTI dataset for (a)3D-CNN without masked-attention (b)3DCMA}~\label{tab:table6}
\end{table}

%\begin{table}[h]
%  \centering
%  \begin{tabular}{l r r r r}
%    % \toprule
%    & & \multicolumn{2}{c}{\small{\textbf{Evaluation %Metric}}} \\
%    \cmidrule(r){2-5}
%    {\small\textit{}}
%    & {\small \textit{RMSE}}
%      & {\small \textit{MAE}}
%    & {\small \textit{RMSE}}
%    & {\small \textit{MAE}}\\
%     {\small\textit{Method}}
%    & {\small \textit{KITTI}}
%      & {\small \textit{KITTI}}
%    & {\small \textit{nuImages}}
%    & {\small \textit{nuImages}}\\
%    \midrule
%    ViVIT & 5.024 & 4.324 & 1.782 & 1.326\\
%    3D-CNN without MA & 4.437 & 3.617 & 1.698 & 1.315 \\
%    \textbf{3D-CNN with MA} & \textbf{3.290} & %\textbf{2.528} & \textbf{1.297} & \textbf{0.974} \\
%    % \bottomrule
%  \end{tabular}
%  \caption{Evaluation on KITTI and nuImages datasets for (a) ViVIT (b) 3D-CNN without Masked Attention (c) 3D-CNN with Masked Attention}~\label{tab:table1}
%\end{table}

\subsubsection{Cross-Dataset Evaluation}

To take into consideration the generalization ability of the AI models, we conduct evaluations across data sets and report their accuracy. It is to be noted that there is a shift in the domain when testing nuImages-trained AI models on the KITTI dataset due to the reasons stated in section~\ref{section:experiments}. To test KITTI models on the nuImages dataset, we need ten frames within a duration of 1 [sec] from nuImages. Since the FPS of the nuImages dataset is only 2 FPS, we are unable to encapsulate ten frames within a temporal window of [1sec]. For this reason, we discard testing KITTI models on the nuImages dataset. We pre-process the KITTI video stream to evaluate nuImages-trained models on the KITTI dataset to ensure the temporal windows are compatible. nuImages-trained models require the temporal window to be 13 frames across 6 [secs]. However, KITTI dataset video streams are sampled at 10 Hz. We apply frame decimation to sample the video at 2 Hz and concatenate frames across 6 [secs] of the stream to encapsulate the $13$ frames temporal window. We intentionally do not resize the images and allow the mismatch in the image dimensions between the two datasets to diversify the gap between them in our evaluation. We report the results for two models below in Table~\ref{tab:table2}.

\begin{table}[h]
  \centering
  \begin{tabular}{l r r}
    % \toprule
    & \multicolumn{2}{c}{\small{\textbf{Evaluation Metric}}} \\
    \cmidrule(r){2-3}
    {\small\textit{Method}}
    & {\small \textit{RMSE (KITTI)}}
      & {\small \textit{MAE (KITTI) }}\\
    \midrule
    ViViT \textit{(nuImages)} & 7.420 & 5.957\\
    \textbf{3DCMA} \textit{(nuImages)} & \textbf{5.880} & \textbf{4.694}\\
    % \bottomrule
  \end{tabular}
  \caption{Evaluation of nuImages trained models on KITTI test data for (a) ViViT (b) 3DCMA}~\label{tab:table2}
\end{table}

\section{DISCUSSION}

In this paper, we propose a modified 3D-CNN architecture with masked-attention employed for ego vehicle speed estimation using single-camera video streams. 3D-CNN is effective in capturing temporal elements within an image sequence. However, we noticed that presence of background clutter and non-cohesive motion within the video stream often confused the model. To extend some control over the focus regions within the images, we proposed to employ a masked-attention mechanism to steer the model to focus on relevant regions. We concatenated the lane segmentation mask as an additional channel to the input images before feeding them to the 3D-CNN. We were able to demonstrate better performance in our evaluations with the inclusion of the proposed masked-attention.

We evaluated the performance of our proposed architecture on two publicly available datasets - nuImages and KITTI. Though there are prior works utilizing the KITTI dataset for the ego vehicle speed estimation task, none clearly stated the train and test splits being used for reporting the results. In this paper, we report the train and test splits from KITTI Road and City categories and hope this will encourage further benchmarks in the future incorporating the same train and test splits. With respect to the nuImages dataset, we believe that we are the first to incorporate the dataset for the task of ego-vehicle speed estimation.

In terms of evaluation, we compared our proposed method against a recent state-of-the-art transformer network for videos, ViViT. We additionally investigated the impact of employing masked-attention to 3D-CNN and saw that the injection of masked-attention improved the MAE and RMSE scores across all scenarios. The increase in the RMSE and MAE scores for cross-dataset evaluation is expected due to the domain gap between the two datasets. However, 3DCMA continued to perform better for the cross-data set evaluation as well. 

One limiting factor we noticed in these datasets is the lack of driving data for higher vehicle speeds. The vehicle speeds are available only up to $20 m/s$, thus limiting the scope of deploying these models for highway driving scenarios.

\section{CONCLUSION}
Though ViViTs can model long-range interactions across videos right from the first layer, we demonstrated that a 3D-CNN injected with masked-attention performed better overall across all test scenarios. In this paper, we introduced a simple yet effective 3D-CNN with masked-attention architecture that can effectively compute the ego-vehicle speed using monocular camera streams. Immediate future work is the extension of current work to utilize the speed of ego vehicle to estimate the speeds and locations of environment vehicles for in-vehicle motion and path planning.

%%%%%%%%%%%%%%%%%%%%%%%%%%%%%%%%%%%%%%%%%%%%%%%%%%%%%%%%%%%%

\appendix
\bibliographystyle{unsrt}
\bibliography{speed}

\end{document}